\documentclass[a4paper,fleqn]{cas-dc}
\usepackage{cite}
\usepackage[numbers,sort&compress]{natbib}
\usepackage{bm}
\usepackage{multirow}
\usepackage[normalem]{ulem}
\useunder{\uline}{\ul}{}
\def\tsc#1{\csdef{#1}{\textsc{\lowercase{#1}}\xspace}}
\tsc{WGM}
\tsc{QE}
\tsc{EP}
\tsc{PMS}
\tsc{BEC}
\tsc{DE}
\begin{document}
\let\WriteBookmarks\relax
\def\floatpagepagefraction{1}
\def\textpagefraction{.001}

\shorttitle{Video Anomaly Detection}

\shortauthors{Yang Wang et~al.}

\title [mode = title]{A Lightweight Video Anomaly Detection Model with Weak Supervision and Adaptive Instance Selection}

%\tnotetext[1]{United National Key Fund}

\author[1]{Yang Wang}[role=,style=Chinese,orcid=0000-0001-7511-2910]
\ead{tongji_wangyang@tongji.edu.cn}

\credit{Writing—original draft, Conceptualization, Methodology, Formal analysis}

\author[2]{Jiaogen Zhou}[style=Chinese]
\ead{zhoujg@hytc.edu.cn}
\credit{Review, editing, Conceptualization}

%\author[3]{Shuigeng Zhou}[style=chinese]
%\ead{sgzhou@fudan.edu.cn}
%\credit{Big data, Deep learning}

\author[1]{Jihong Guan}[role=Corresponding author,
   style=Chinese
   ]
\cormark[1]
\ead{jhguan@tongji.edu.cn}
%\ead[URL]{www.sayahna.org}
\credit{Conceptualization, Resources, Review, editing}

\cortext[cor1]{Corresponding author}
%\cortext[cor2]{Principal corresponding author}

\address[1]{Tongji University, No.~4800 Cao'an Highway, Shanghai, 201804, China}
\address[2]{Jiangsu Province Engineering Research Center for Intelligent Monitoring and Management of Small Water Bodies, Huaiyin Normal University, Huaian 223300, China}
%\address[3]{Shanghai Key Lab of Intelligent Information Processing, and the School of Computer Science, Fudan University,Shanghai, 200438, China}

\begin{abstract}
Video anomaly detection is to determine whether there are any abnormal events, behaviors or objects in a given video, which enables effective and intelligent public safety management. As video anomaly labeling is both time-consuming and expensive, most existing works employ unsupervised or weakly supervised learning methods. This paper focuses on weakly supervised video anomaly detection, in which the training videos are labeled whether or not they contain any anomalies, but lack information about the specific frames and quantities of anomalies. However, the uncertainty of weakly labeled data and the large model size prevent existing methods from wide deployment in real scenarios, especially the resource-limit situations such as edge-computing. In this paper, we develop a lightweight video anomaly detection model. On the one hand, we propose an adaptive instance selection strategy, which is based on the model's current status to select confident instances, thereby mitigating the uncertainty of weakly labeled data and subsequently promoting the model's performance. On the other hand, we design a lightweight multi-level temporal correlation attention module and an hourglass-shaped fully connected layer to construct the model, which can reduce the model parameters to only 0.56\% of the existing methods (e.g. RTFM). Our extensive experiments on two public datasets UCF-Crime and ShanghaiTech show that our model can achieve comparable or even superior AUC score compared to the state-of-the-art methods, with a significantly reduced number of model parameters. 

%Experimental results on real datasets show that our proposed method significantly improves the performance of video anomaly detection tasks and has the best overall performance in terms of effectiveness and efficiency metrics.

\end{abstract}

% \includegraphics{grabs.pdf}

\begin{highlights}
\item A new Lightweight video anomaly detection model is proposed .
\item Weakly labeled data  problem is mitigated  by an adaptive sampling strategy .
\item A lightweight multi-level temporal correlation attention module is designed.
\item A lightweight hourglass-shaped fully connected layer is designed.
\item Extensive experiments have shown that the proposed method is both lightweight and effective.
\end{highlights}

% Keywords

\begin{keywords}
\sep Video Anomaly Detection
\sep Weak Supervision \sep Adaptive Instance Selection \sep Lightweight Model
\end{keywords}

\maketitle

\section{Introduction}
surveillance serves as a critical tool for identifying unexpected or abnormal events in many scenarios such as traffic monitoring and public safety management. Traditionally, video surveillance is heavily dependent on manual operations and of low intelligence~\cite{DBLP:journals/tsmc/SodemannRB12, DBLP:journals/tsmc/qq,DBLP:conf/cvpr/SultaniCS18}. For example, many cameras are installed in public venues such as stations and parks, to monitor unexpected or abnormal events, which generate huge amounts of videos. To check these videos manually is time-consuming and laborious. The rapid development of computer vision and deep learning technologies has spurred more and more research on video abnormal event detection or video anomaly detection (VAD), which enables the applications of automatic scene monitoring and intelligent early warning.

Generally, there are there types of video anomaly detection methods, supervided~\cite{sapkota2022bayesian}, unsupervised~\cite{ref6,DBLP:conf/cvpr/PangYSH020,ref7,ref8,ref9} and weakly-supervised~\cite{DBLP:conf/cvpr/SultaniCS18,DBLP:conf/icip/ZhangQM19,DBLP:conf/eccv/Wu0SSSWY20,DBLP:conf/icmcs/WanFXM20,DBLP:conf/iccv/TianPCSVC21,DBLP:journals/tmm/ChangLSFZ22,watanabe2022real}. Supervised video anomaly detection typically requires frame-level or even pixel-level labels, which incurs expensive training cost. Hence, there is a little related research in this direction. Unsupervised video anomaly detection uses unlabeled data to train models, with the lowest training cost, but exhibits poor  performance. Weakly-supervised video anomaly detection (WVAD) uses weakly-labeled data to train the model, where the training videos are labeled whether or not they contain any anomalies, but there
is no information about which frames the anomalies are located. Thus, WVAD
inherits the advantages of supervised and unsupervised methods. On the one hand, WVAD has better performance than unsupervised methods as some supervision is exploited. On the other hand, WVAD is much cheaper and more efficient in acquiring the training data than supervised methods, because the latter requires to label each frame whether it contains anomalies. Therefore, WVAD becomes a hot topic of video anomaly detection.

%is not only accurate and facilitates obtaining annotated data, but it is also one of the video anomaly detection methods widely studied at present. 
%According to the data characteristics,
WVAD is typically based on Multiple Instance Learning (MIL)~\cite{DBLP:conf/cvpr/SultaniCS18,DBLP:conf/icip/ZhangQM19}. Under the MIL framework, a video is viewed as a bag that consists of various clips, each of which is considered as an instance. For the training videos, the annotations are on the video level. That is, we know which videos have anomalies, but we do not know which clips (or instances) and frames have anomalies. In this context, WVAD methods face two major challenges. The first challenge is the uncertainty of the weakly labeled data: we do not know both the number and the locations of anomalous clips in each anomalous video, which limits the full exploitation of the training anomalous data, thus resulting in unsatisfactory performance. The second challenge is the huge model size. The models of existing methods have too many parameters, making them difficult to be applied in resource-pressing scenarios, such as edge-computing applications. Existing methods have been mainly trying to tackle these two challenges.

\begin{figure*}
	\centering
		\includegraphics[scale=0.55]{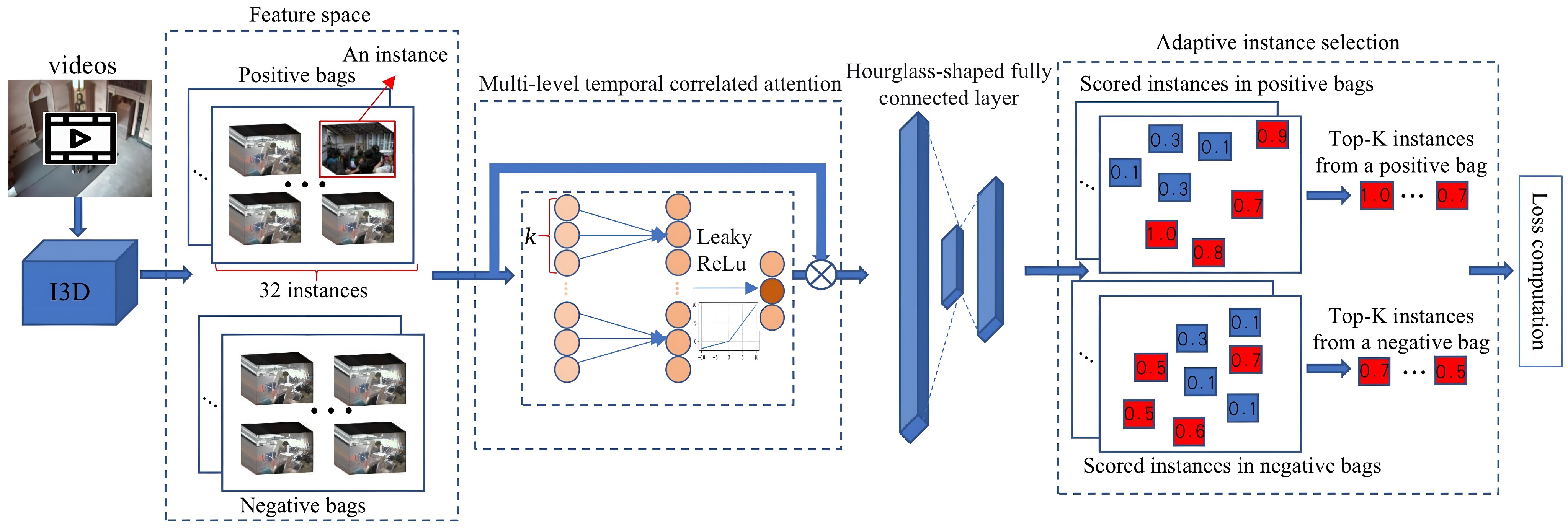}
	\caption{The framework of our method Light-WVAD. Our model is based on the multi-instance learning (MIL) framework. Each video is divided into 32 consecutive clips (or instances), which are grouped into a positive instance bag (for abnormal videos) or a negative instance bag (for normal videos). Video features are extracted by I3D. A \textit{Multi-level Temporal correlation Attention} (MTA) module is designed to capture time-related information, which is then input to a \textit{Hourglass-shaped Fully Connected layer} (HFC) to calculate the score of each instance. The top-$K$ reliable instances are selected based on an \textit{Adaptive Instance Selection} (AIS) strategy for subsequent loss calculation.}
	\label{FIG:2}
\end{figure*}

Concretely, existing MIL based methods~\cite{DBLP:conf/cvpr/SultaniCS18,DBLP:conf/eccv/Wu0SSSWY20, DBLP:conf/icip/ZhangQM19,DBLP:conf/icmcs/WanFXM20,ref16} adopt a strategy to maintain balance between the number of anomalous videos and that of normal videos in the training set, and compute the loss by selecting the instance (or clip) of the highest anomaly score. This strategy aims to minimize the uncertainty of weakly labeled data. Typically, in real scenarios, the time duration of anomalous videos used for training usually exceeds 30 seconds, which suggests the presence of multiple anomalous clips. However, the selection of only the instance (clip) of the highest anomaly score might lead to under-utilization of all the available anomalous data. Statistical finding from RTFM~\cite{DBLP:conf/iccv/TianPCSVC21} reveals that the mean number of anomalous clips within an anomalous video is approximately three. Therefore, choosing the three instances (clips) of the largest feature magnitude in both anomalous and normal videos for loss calculation can more effectively utilize the training data, thereby improving performance. However, our point of view is that different numbers of instances should be chosen for different datasets. Therefore, we propose an \textit{Adaptive Instance Selection} (AIS) strategy, which selects the number of normal or anomalous instances adaptively based on the current status of the model in training, thereby it is able to accommodate different training datasets.

On the other hand, there are relatively fewer lightweight models for VAD in the literature~\cite{DBLP:journals/tmm/ChangLSFZ22,watanabe2022real,ref16}, which usually compromise performance to embrace smaller models. In this paper, to devise a lightweight yet high-performance anomaly detection model, we develop a lightweight \textit{Multi-level Temporal correlation Attention} (MTA) module. This module emphasizes the relationships between video instances (clips) of different time spans, thus making the model focus on the important instances in the videos. Furthermore, we design an \textit{Hourglass-shaped Fully Connected layer} (HFC), which contains only half the parameters of a conventional fully connected layer (FC), yet does not degrade the model's performance. Finally, Observing the limitation of the sparsity loss widely used in existing methods, we specifically introduce a more effective antagonistic loss.

In summary, in this paper we propose a novel, MIL-based lightweight video anomaly detection model, whose framework is shown in Fig.~\ref{FIG:2}. First, we use I3D to decompose each video into 32 continuous clips, which are encoded and organized into the positive or negative instance bag. The instances in the bags are input into the Multi-level Temporal correlation Attention (MTA) module, to make the model focus on the important features in the time dimension of the videos. Next, instance features are input into an Hourglass-shaped Fully Connected (HFC) layer to obtain the anomaly score for each instance. Finally, we dynamically determine the number $K$ of positive and negative instances used for loss calculation based on the adaptive instance selection strategy. The main contributions of this study include:

\begin{enumerate}
\itemsep=0pt
\item We propose an adaptive sample selection strategy that alleviates the uncertainty problem of weakly-labeled data and improves model performance.
\item We design a lightweight multi-level temporal correlation attention module and an hourglass-shaped fully connected layer, which leads to a lightweight video anomaly detection method with only 0.56\% of the parameters of the existing SOTA method RTFM~\cite{DBLP:conf/iccv/TianPCSVC21}.
\item We analyze the limitations of using sparsity loss in weakly-supervised video anomaly detection, and develop a more suitable antagonistic loss for this problem.
\item We conduct extensive experiments on two benchmark datasets UCF-Crime and ShanghaiTech, which show that our model achieves competitive or even superior performance compared to state-of-the-art methods of video anomaly detection.
\end{enumerate}

The rest of this paper is organized as follows: Sec.~\ref{sec:related-work} reviews the related works and highlights the differences of our work from the existing ones. Sec.~\ref{sec:method} introduces the proposed method in detail. Sec.~\ref{sec:performance} presents the results of performance evaluation, including performance comparison with existing works and ablation studies. Finally, Sec.~\ref{sec:conclusion} concludes the paper and pinpoints the future works.

\section{Related Work}\label{sec:related-work}
Here we review the related work from two aspects: weakly supervised video anomaly detection and attention network.

\subsection{Weakly Supervised Video Anomaly Detection}
Traditional anomaly detection methods typically assume that only normal training data is available and use hand-crafted features for one-class classification to solve the problem~\cite{DBLP:journals/pami/MedioniCBHN01,DBLP:conf/cvpr/BasharatGS08,DBLP:conf/cvpr/ZhangLL09,DBLP:conf/cvpr/WangSLRWPCW14}. With the development of deep learning techniques, some unsupervised learning methods utilize deep neural networks to learn features such as human posture and optical flow~\cite{ref6,ref20,ref21,ref22,DBLP:journals/tnn/FangLZXY22}, or utilize the difference in feature distribution between normal and abnormal samples~\cite{ref9} to perform abnormality detection. The essence of unsupervised video anomaly detection methods~\cite{ref6,ref20,ref21,ref22,DBLP:journals/tnn/FangLZXY22} lies in the assumption that anomalies are rare events, and therefore the model is trained to learn mainly the feature distribution of normal samples. Based on this assumption, the model then determines whether a test sample is an anomaly by calculating the reconstruction errors or feature distributions of the test sample and the normal samples. However, due to the lack of prior knowledge about anomalies, these methods are prone to overfitting to training data and unable to distinguish normal and anomalous events.

Some works~\cite{DBLP:conf/cvpr/SultaniCS18,DBLP:conf/eccv/Wu0SSSWY20,DBLP:conf/icip/ZhangQM19,DBLP:conf/icmcs/WanFXM20,DBLP:conf/iccv/TianPCSVC21,DBLP:conf/eccv/HongHLZ20,DBLP:journals/spl/ZaheerMSL20} have shown that utilizing partially labeled abnormal samples can achieve better performance than unsupervised methods. However, the cost of obtaining a large number of frame-level labels is prohibitively high. Therefore, some video anomaly detection methods apply video-level labels for weakly supervised training. Sultani et al.~\cite{DBLP:conf/cvpr/SultaniCS18} proposed a method that uses video-level labels and introduced the large-scale \textit{weakly supervised video anomaly detection} (WVAD) dataset UCF-Crime. This makes WVAD one of the mainstream research directions in video anomaly detection~\cite{DBLP:conf/icip/ZhangQM19,DBLP:conf/iccv/TianPCSVC21,DBLP:conf/aaai/00070J22,DBLP:journals/spl/ZhangHLX22}.

Existing Weakly supervised video anomaly detection methods are mainly based on multple-instance learning (MIL). Given that the training data possesses only video-level labels, these approaches typically employ the instances of the highest anomaly prediction score from both positive and negative bags for loss computation during training, leading to under-utilization of the training data. To address this issue, Zhong et al.~\cite{DBLP:conf/cvpr/ZhongLKLLL19} transformed WVAD into a binary classification problem in the presence of label noise. They employed a \textit{Graph Convolutional Neural Network} (GCN)~\cite{DBLP:conf/aaai/LiWZH18} to eliminate label noise, thus enhancing data utilization and model performance. While this method improves model performance, the training computation cost associated with GCN and MIL is considerably high. Furthermore, it may make features unconstrained in the feature space, resulting in unstable performance. Furthermore, Tian et al.~\cite{DBLP:conf/iccv/TianPCSVC21} amalgamated representation learning and anomaly score learning by devising \textit{Robust Temporal Feature Magnitude} (RTFM) learning. They separately selected the three instances of the largest  feature magnitude, in both abnormal and normal bags for loss calculation. Such an approach can effectively utilize the training data, thus achiveving better performance. However, this training data utilization strategy does not consider the diverse characteristics of different datasets. 

In this paper, we propose an adaptive instance selection strategy, which adaptively selects the numbers of normal and abnormal instances for subsequent optimization, based on the current training status of the model, aiming to boost the utilization of weak label datasets.

\subsection{Attention Network}
Attention networks are initially used for machine translation~\cite{DBLP:conf/nips/VaswaniSPUJGKP17} and later widely used for various computer vision tasks such as image classification~\cite{DBLP:conf/cvpr/WangJQYLZWT17}, object detection~\cite{DBLP:conf/cvpr/CaoCLL20}, image segmentation~\cite{DBLP:conf/cvpr/ChenYWXY16}, image captioning~\cite{DBLP:conf/icml/XuBKCCSZB15}, and action recognition~\cite{DBLP:journals/tmm/WuML20} etc., and have achieved excellent performance. Recently, attention networks have also been employed in weakly supervised tasks. Choe et al.~\cite{DBLP:conf/cvpr/ChoeS19} proposed an attention-oriented dropout layer, leveraging self-attention mechanisms to address the weakly supervised object localization problem. W-TALC~\cite{DBLP:conf/eccv/PaulRR18} combines MIL and common activity similarity loss to train an attention module to solve the weakly supervised action localization problem in videos. Zhou et al.~\cite{ref7} proposed an attention that focuses on the foreground of an image to alleviate the foreground-background imbalance problem in anomaly detection. Li et al.~\cite{DBLP:journals/kbs/LiYXBZ22} improved the ability of weakly supervised anomaly detection models to extract relationships between video frames by using SE-attention~\cite{DBLP:conf/cvpr/HuSS18}. However, this method, not specifically designed for video data, fails to enable the model to concentrate on the temporal correlation between consecutive instances (clips), and has a relatively large number of parameters. 

In this paper, We devise a Multi-level Temporal Correlation Attention module for video data with temporal relationships. This helps the model to focus attention on important instances (clips) in the video. Furthermore, its limited number of parameters makes it more suitable for video anomaly detection models in resource-constrained scenarios.

\begin{figure}[!t]
\centering
\includegraphics[width=3.2in]{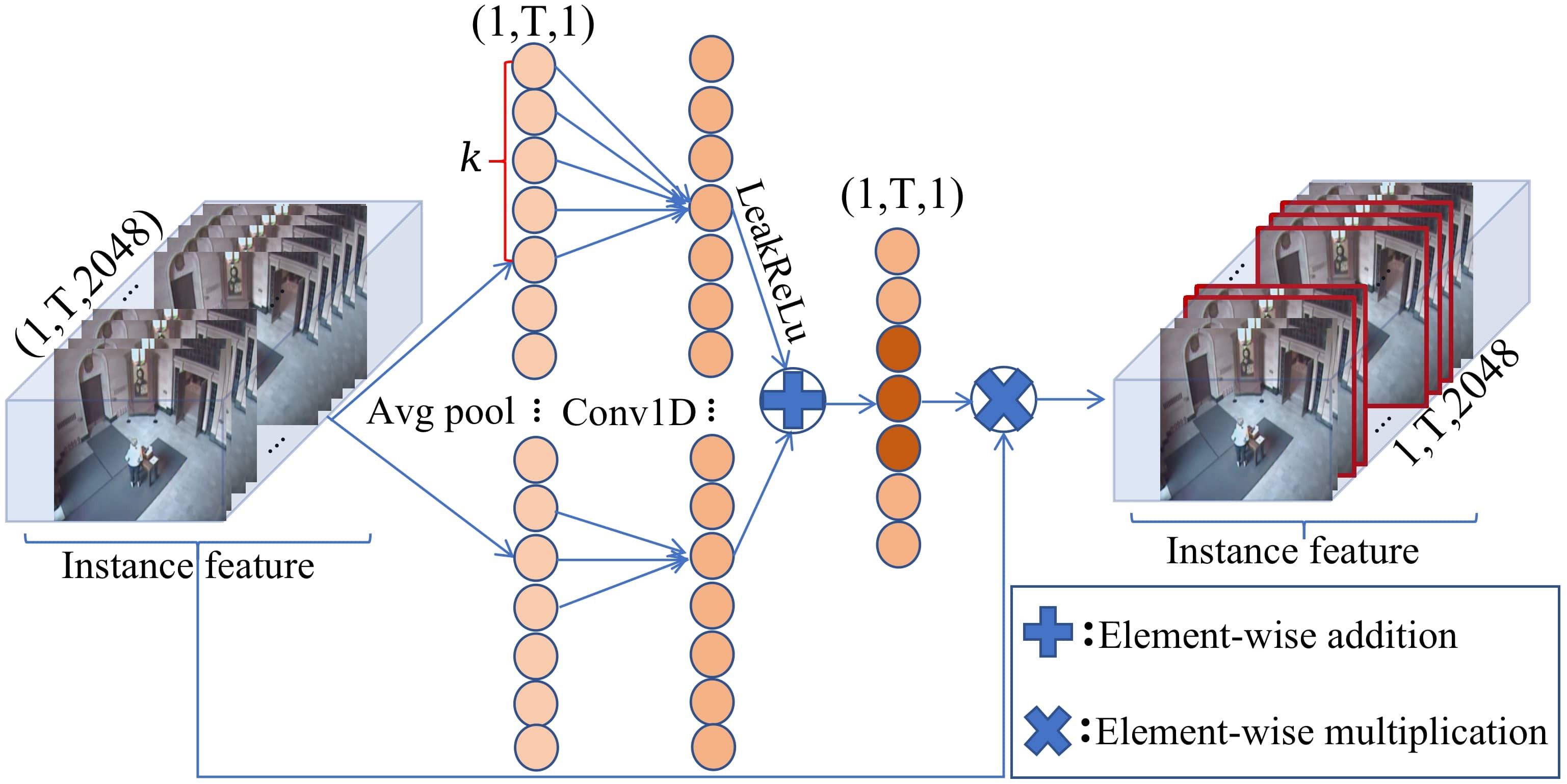}
\caption{The structure of the multi-level temporal correlation attention (MTA) module. Here, each video is divided into $T$ (32 in this paper) clips, each of which corresponds to an instance.}
\label{FIG:1}
\end{figure}

\section{Methodology}\label{sec:method}
This section presents the proposed \textit{Lightweight Weakly Supervised Video Anomaly Detection} (Light-WVAD in short) method. The framework of Light-WVAD is shown in Fig.~\ref{FIG:2}. Light-WVAD mainly comprises a lightweight multi-level temporal correlation attention module, an hourglass-shaped fully connected layer, and an adaptive instance selection strategy to alleviate the uncertainty of weakly labeled data. In addition, we employ a more robust antagonistic loss to further optimize the model's performance. In what follows, we first give a formal definition of the problem, then introduce the major modules of our method in detail. 
%-------------------------------------------------------------------------
\subsection{Problem Statement}
In the context of weakly-supervised learning, anomaly detection can typically be regarded as a multiple-instance learning problem. Given a set of videos with video-level annotations (i.e., just labeling whether or not a video in the training dataset contains anomalous content),  Weakly-supervised video anomaly detection (WVAD) aims to train a model with the annotated data, which is able to predict whether there are anomalies in any new videos.

\subsection{Feature Extraction}
In the data preprocessing stage, the widely used networks for feature extraction are I3D~\cite{DBLP:conf/cvpr/CarreiraZ17} and C3D~\cite{DBLP:conf/iccv/TranBFTP15}. Some existing research works~\cite{feng2021mist,DBLP:conf/iccv/TianPCSVC21,DBLP:conf/aaai/00070J22} have shown that I3D can more effectively extract sample features. Therefore, in this paper we use I3D for data preprocessing and convert videos into feature vectors. Concretely, Given a video $V_{i}$, we divide it into $32$ consecutive and non-overlapping clips, each of which is regarded as an instance. The clips are grouped into a positive or negative bag based on the video-level labels $Y$. Here, the positive bag ($Y=1$) contains at least one anomalous instance (clip), while the negative bag ($Y=0$) consists only of normal instances (clips).

\subsection{Multi-level Temporal Correlation Attention}\label{sec:mta}

The structure of the Multi-level Temporal Correlation Attention (MTA) module is shown in Fig.~\ref{FIG:1}. In this module, we first use a global average pooling layer to convert the $T$ ($T$ is 32 in this paper) instance features in a bag into a $T$-dimensional vector representing $T$ channels. Then, we evaluate cross-channel interactions via convolution, and ultimately determine the weight of each channel. This makes the model focus on the important instances of each video in the time dimension.

To reduce parameters, we use a one-dimensional convolution $Conv1D$ of kernel size $k$ ($k$ $\geq$ 3) to capture the cross-channel interactions of adjacent instances. In addition, in order to capture the interaction information between adjacent channels at different time spans, we jointly use convolution kernels of different sizes. The calculation process is as follows:
\begin{equation}
\begin{split}
   T_{attention}=(Conv1D[k], Conv1D[k-2], \dots, \\ Conv1D[3]) 
   \ast G(\chi)
   \label{equa-1}
\end{split}
\end{equation}

In Equ.~(\ref{equa-1}), $Conv1D[k]$ represents the convolution kernel function of size $k~(k \leq T)$, \textit{$*$} denotes the convolution operation, $\chi$ is the feature vector of a video and  $G(\chi)$ is the feature vector after global pooling. In the one-dimensional convolution layer, the convolution operation between $Conv1D[k]$ and $G(\chi)$ is equivalent to sliding the convolution kernel, and multiplying it with the feature, then adding the convolution results. The calculation of $G(\chi)$ is as follows:
\begin{equation}
\begin{split}
   G(\chi) = \frac{1}{WH}  {\textstyle \sum_{i=1,j=1}^{W,H}} \chi_{ij}   
   \label{equa-2}
\end{split}
\end{equation}

Finally, by feeding $T_{attention}$ into the activation layer LeakRelu~\cite{DBLP:journals/jmlr/GlorotBB11} and performing a concatenation operation with the original input $G(\chi)$, the attention module outputs $Y_{attention}$:

\begin{equation}
\begin{split}
   Y_{attention}=\lambda_{1}Sum(L\_Relu(T_{attention})) \cdot \chi 
   \label{equa-3}
\end{split}
\end{equation}
Here,  $L\_Relu()$~\cite{DBLP:journals/jmlr/GlorotBB11}is the activation function, $\lambda_{1}$ is a hyperparameter set to 0.1, and $\cdot $ represents the corresponding multiplication operation. $Y_{attention}$ is the video feature with multi-level temporal correlations obtained by MTA, which can enhance the features of important instances in each video and weaken the features of less important instances.

\subsection{Hourglass-shaped Fully Connected Layer}

\begin{figure}[!t]
	\centering
		\includegraphics[scale=0.6]{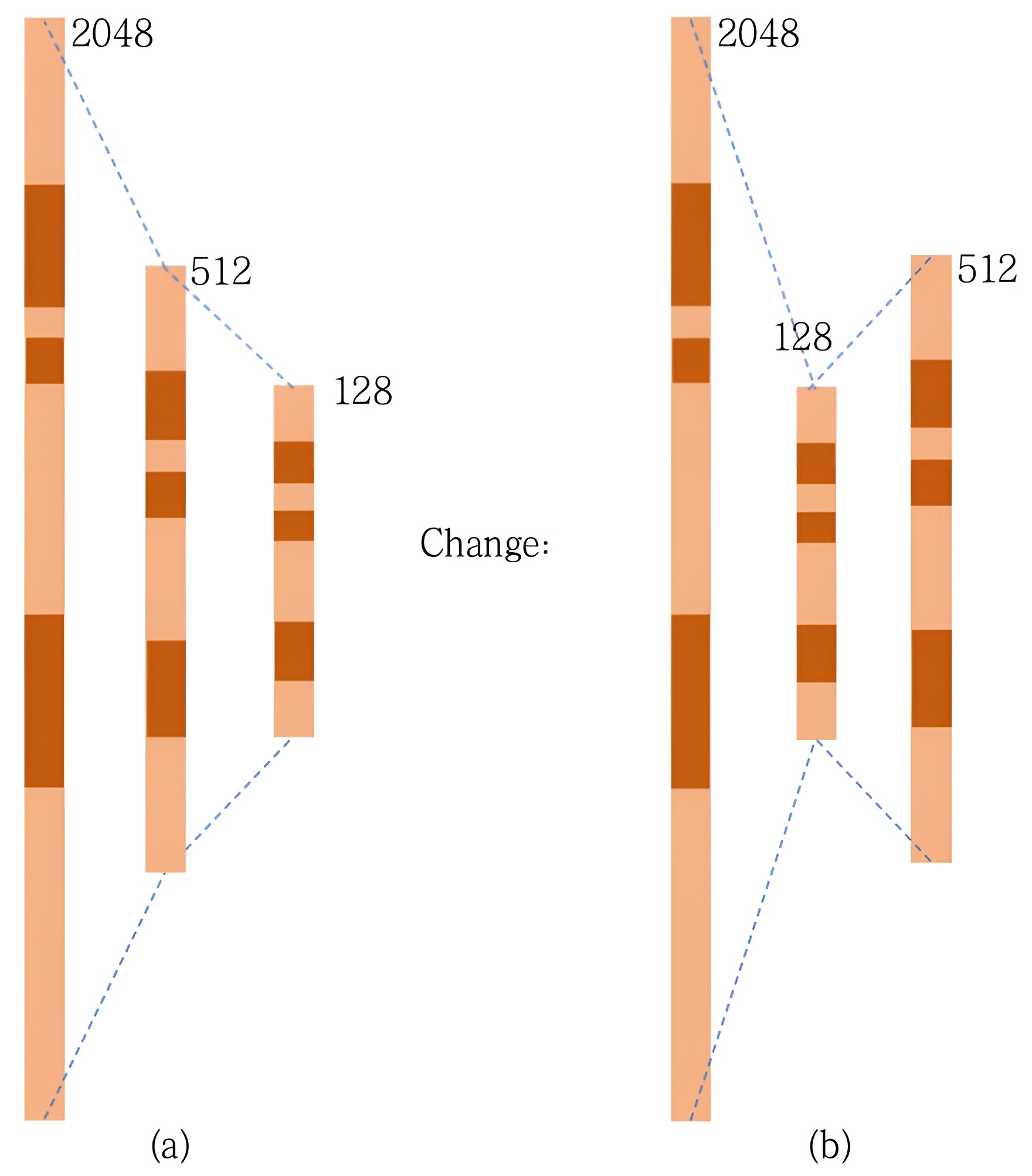}
	\caption{The structures of (a) the traditional fully connected layer (FC), and (b) our hourglass-shaped fully connected layer (HFC).}
	\label{FIG:3}
\end{figure}

In the fully connected layer, each node connects with all the nodes of the preceding layer, resulting in a substantial number of parameters in the fully connected layer. To cut off the parameters in this layer, we propose a novel \textit{hourglass-shaped fully connected} (HFC) layer. Fig.~\ref{FIG:3} illustrates the HFC structure on the right side, for comparison the traditional fully connected (FC) layer is illustrated on the left. HFC and FC have the same number of layers and layer dimensions, however, HFC has a \textit{2048-64-128} structure, in contrast to the traditional FC's \textit{2048-128-64}. Consequently, the number of parameters in HFC is approximately half of that in FC. With HFC, we get the anomaly scores of all instances in the positive and negative bags.

\subsection{Adaptive Instance Selection}
Here, we design an adaptive instance selection (AIS) strategy to choose important instances based on the current training status of the model and autonomously determines the number of instances used for loss computation. This strategy is primarily based on two important facts of MIL-based weakly supervised video anomaly detection. First, the negative bag contains only normal instances, i.e., each instance in the negative bag is normal. Second, as continuity exists amongst instances in the videos, the predicted anomaly scores of neighboring instances should also be continuous. These two factors enable us to infer the model's training status by examining its predictions of negative instances during the training process. With this, the confident positive (anomalous) instances can be determined from the set of current positive instances.

\begin{figure*}
\begin{center}
	\centering
\includegraphics[scale=0.6]{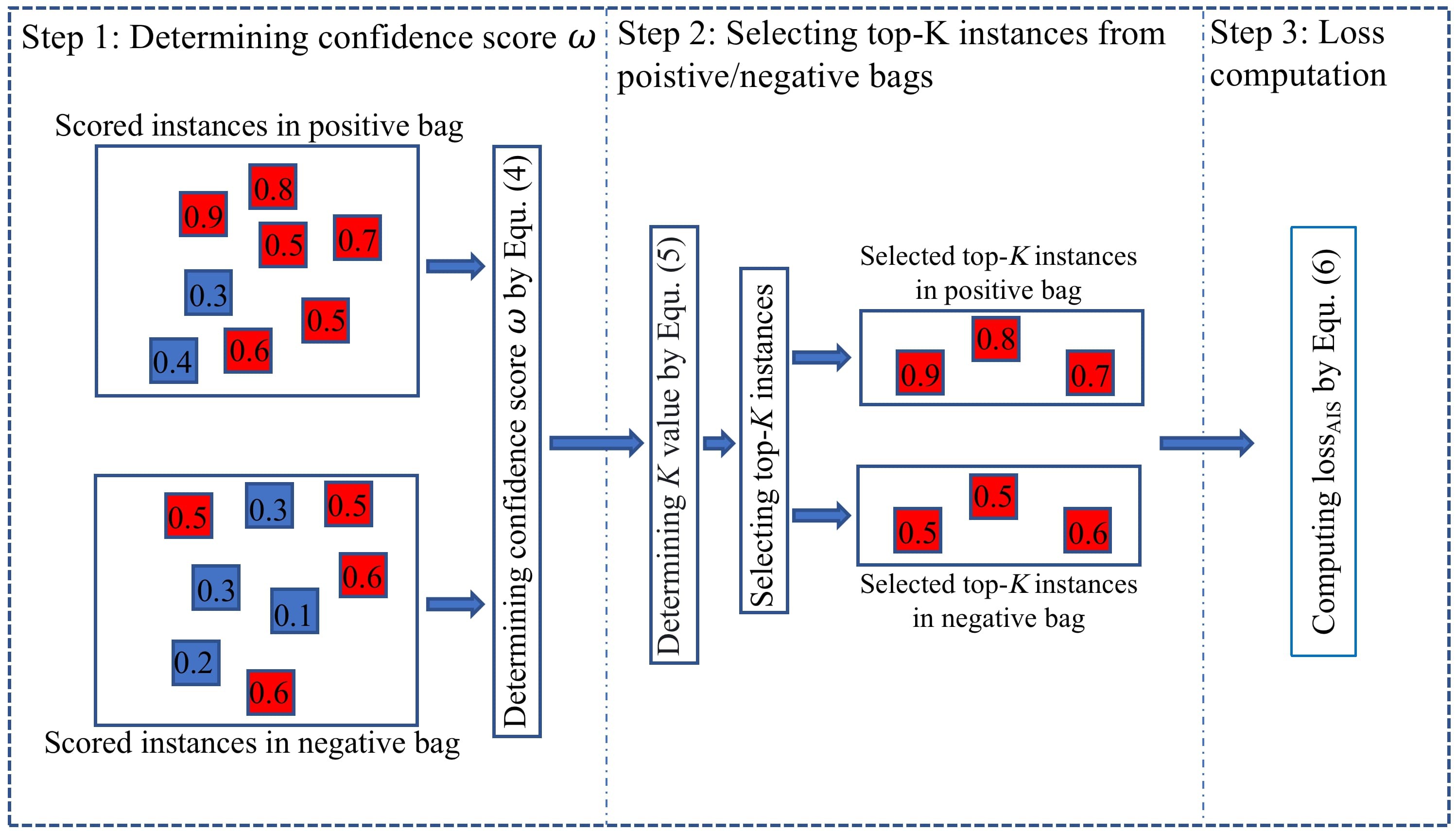}
	   \caption{The workflow of our adaptive instance selection (AIS) strategy on a pair of positive and negative bags. It consists of three steps. Here, each red or blue square is an instance. Top-$K$ instances are selected from both the positive and negative bags.}
	\label{FIG:4}
\end{center}
\end{figure*}

The workfolow of AIS is shown in Fig.~\ref{FIG:4}, which consists of three steps: 

\textbf{Step 1.} With the anomaly scores of all instances from the HFC module, the first step of AIS is to calculate the confidence score $\omega$ that measures the mature degree of the model (i.e., how well the model is trained), based on the average anomaly scores of the negative instances and the average score difference between consecutive positive and negative instances. Concretely, 

\begin{equation}
\begin{split}
   \omega = 1-\frac{1}{T} {\textstyle \sum_{i=1}^{T}} S_{i}^{N} -\frac{1}{2T-2} {\textstyle \sum_{i=1}^{T-1}} (\left | S_{i+1}^{N}-S_{i}^{N}\right |  \\
   +\left | S_{i+1}^{P}-S_{i}^{P}\right |)
   \label{equa-4}
\end{split}
\end{equation}
Above, $T$ is the number of instances in each positive/negative bag, $S^P$ and $S^N$ are the scores of positive and negative instances respectively. In Equ.~(\ref{equa-4}), the 2nd item computes the mean of anomaly scores of negative instances, while the 3rd item calculates the mean of the differences in anomaly scores between consecutive positive and negative instances. As training goes, the 2nd term approaches 0 as negative instances should have 0 anomaly score, and the 3rd term also be close to 0 considering the continuity of anomaly scores of consecutive instances. Thus, the confidence score $\omega$ becomes close to 1. 

\textbf{Step 2.} We select confident instances for loss computation based on the confidence score $\omega$. The number of confident instances to be selected is determined as follows:

\begin{equation}
    K=\omega * {\textstyle \sum_{i=1}^{T}}f(S_{i}^{P}) 
    \label{equa-5}
\end{equation}

where the function $f(.)$ is 1 if $S_{i}^{P} \geq 0.9$, otherwise is 0. That is, we first count the number of instances with score over 0.9, and then multiply it by the confidence score $\omega$ to obtain the final number of confident anomaly instances $K$. 

\textbf{Step 3.}The feature magnitudes~\cite{DBLP:conf/iccv/TianPCSVC21} of the top-$K$ instances with the highest scores are selected separately from the positive and negative sets, and corresponding optimizations are performed accordingly. Thus, the loss function is as follows:
\begin{equation}
\begin{split}
    loss_{AIS}=\sum_{\chi\in U_{\chi\_top-K} } (ylog(mean(s_{\theta }(\chi))) \\
    +(1-y)log(1-mean(s_{\theta }(\chi))))
     \label{equa-6}
\end{split}
\end{equation}
where $s_{\theta }$ is the feature extraction part of the model, $mean$ is the mean function. $U$ is the set of video instance (clip) features extracted by I3D, and $U_{\chi\_top-K}$ is the set of features of the top-$K$ instances selected from the positive and negative bags, $\chi$ is the feature of an instance.  $y\in\{0,1\}$, and $y=0$ when $\chi$ is a normal video clip feature, otherwise $y=1$.

\subsection{Antagonistic Loss Function}
Existing methods of weakly supervised anomaly detection predominantly employ smooth loss and sparsity loss for model optimization. Smooth loss assumes that there exists only slight discrepancy in features between consecutive instances, and thus the variation in anomaly scores is tiny, which conforms to the real scenario. On the other hand, the sparsity loss assumes that anomaly events in videos occur infrequently, therefore the mean anomaly score of instances in an anomalous video should approach zero. Although the assumption of sparsity is reasonable, it does not hold in actual model training. Owing to computational resource constraint, typically 32 consecutive instances are used to form a bag. In training, when anomalies are present in a bag, the proportion of abnormal instances cannot be overlooked. According to our preliminary analysis conducted on ShanghaiTech~\cite{DBLP:conf/cvpr/ZhongLKLLL19}, the ratio of abnormal instances approximates 20\%, and the mean score is around 0.2. Based on the above analysis, the use of smoothing loss is retained in this study. The implementation of smoothing loss is defined by the following formula:

\begin{equation}
loss_{smooth}=\frac{1}{T-1} {\textstyle \sum_{i=1}^{T-1}} (\left \| S_{i+1}-S_{i}\right \|^2)
   \label{equa-7}
\end{equation}
where $S$ represents the anomaly scores predicted by the model for instances, and \textit{$\left \|  \right \|^2$} is the \textit{$L2$} norm. The $L2$ norm assigns larger losses and gradients to anomalous instances. 

Besides the smooth loss, we also devise an antagonistic loss, which is more suitable for weakly supervised MIL. The antagonistic loss is based on the antagonistic assumption of scores for positive and negative instances, capable of gauging the model's predictive performance on both positive and negative instances. As the negative bag exclusively contains normal instances, the scores of the most normal instances should approach zero. On the contrary, for the positive bag that includes some abnormal instances, the scores of the most anomalous instances should be close to 1. With this in mind, we have the following antagonistic loss:

\begin{equation}
\begin{split}
    loss_{antagonistic}= S_{top-1}^{N}+(1-S_{top-1}^{P})
    \label{equa-8}
\end{split}
\end{equation}

Here, \textit{$S_{top-1}^{P}$} and \textit{$S_{top-1}^{N}$}  respectively represent the highest (top-1) abnormality score of instances in the positive and negative bags predicted by the model. $loss_{antagonistic}$ not only embodies the antagonistic constraint between positive and negative instances, but also provides optimization constraints for them respectively. Finally, the total loss function of our method Light-WVAD is as follows:
\begin{equation}
loss_{all}=loss_{AIS}+loss_{smooth}+loss_{antagonistic}
   \label{equa-9}
\end{equation}

\section{Performance Evaluation}\label{sec:performance}
\subsection{Datasets and Evaluation Metrics}
We evaluate our model on two commonly used video anomaly detection benchmark datasets, ShanghaiTech~\cite{DBLP:conf/cvpr/LiuLLG18} and UCF-Crime~\cite{DBLP:conf/cvpr/SultaniCS18}. The videos in both datasets were collected by fixed recording devices (surveillance cameras).
\begin{enumerate}
\item ShanghaiTech~\cite{DBLP:conf/cvpr/LiuLLG18}: This dataset consists of 437 street surveillance videos captured from fixed angles, with 13 different background scenes. It contains 307 normal videos and 130 abnormal videos. Originally, it was used as a benchmark for unsupervised video anomaly detection. However, Zhong et al.~\cite{DBLP:conf/cvpr/ZhongLKLLL19} reorganized the dataset by selecting a subset of anomalous testing videos to create a weakly supervised training set, so that all 13 background scenes are covered in both the training and testing sets. We follow exactly the same procedure as in Zhong et al.~\cite{DBLP:conf/cvpr/ZhongLKLLL19} to convert ShanghaiTech for the weakly supervised setting.
\item UCF-Crime~\cite{DBLP:conf/cvpr/SultaniCS18}: The UCF-Crime dataset is a real-world surveillance video dataset that consists of 950 anomalous videos belonging to 13 anomalous categories, and 950 normal videos. The training set provides only video-level labels, while the testing set provides both video-level labels and frame-wise annotations for evaluation. Each anomalous video in the testing set contains one or two anomalous events. Due to significant variations in the time duration of abnormal events across different videos, this dataset poses a serious challenge for video anomaly detection models.
\end{enumerate}

For performance evaluation, we adopt the frame-based \textit{receiver operating characteristic} (ROC) curve and the corresponding \textit{area under the curve} (AUC) as performance metrics, which is consistent with that  used in previous studies~\cite{DBLP:conf/cvpr/SultaniCS18,DBLP:journals/tmm/ChangLSFZ22,watanabe2022real,DBLP:journals/tifs/LiuZLK23,wang2023video}.

\subsection{Compared Methods}
We compare our method with a number of existing works, including thirteen general (non-lightweight) models and there lightweight methods. For some models, we consider two versions of using different feature extractors.

Among the thirteen non-lightweight models, the work of Sultani et al.~\cite{DBLP:conf/cvpr/SultaniCS18} is the first to use the MIL framework and weakly supervision for video anomaly detection, and most of the subsequent works follow their settings. The method of Zhang et al.~\cite{DBLP:conf/icip/ZhangQM19} defines a new inner-bag loss in the MIL framework to limit the function space and expand the differences in feature distributions for different types of instances. ARNet~\cite{DBLP:conf/icmcs/WanFXM20} uses a dynamic multi-instance learning loss to enlarge the interclass distance between anomalous and normal instances, and a center loss to narrow the intraclass distance of normal instances to boost the model's ability to distinguish anomalies. CLAWS Net+~\cite{zaheer2022clustering} employs a clustering loss to mitigate labeling noise, which improves the learning of representations for both abnormal and normal videos. MIST~\cite{feng2021mist} adopts a multi-instance self-training framework to effectively refine task-specific discriminative representations using only video-level annotations. Similarly, MSL~\cite{DBLP:conf/aaai/00070J22} proposes a self-training strategy and designs a transformer-based Multi-Sequence Learning network to further learn video-level anomaly probabilities and clip-level anomaly scores. RTFM~\cite{DBLP:conf/iccv/TianPCSVC21} trains the feature magnitude learning function to effectively identify positive instances and improves the robustness of the model. BN-SVP~\cite{sapkota2022bayesian} introduces a novel approach called Bayesian non-parametric submodular video partitioning to enhance the training of MIL models, and provides a reliable solution for robust anomaly detection in practical scenarios that involve outlier instances or multiple types of abnormal events.  DAR~\cite{DBLP:journals/tifs/LiuZLK23} and the method of Wu et al.~\cite{DBLP:conf/eccv/Wu0SSSWY20} design the model structure using a multi-branching and multi-stage approach, respectively, to improve the model's ability to understand and fuse different modal data in the videos. Mu et al.'s method ~\cite{DBLP:journals/ipm/MuSWC22} and GCN-Anomaly ~\cite{DBLP:conf/cvpr/ZhongLKLLL19} innovatively employ graph convolutional networks for enhancing the model's ability to understand spatio-temporal correlation features in videos. The NTCN-ML~\cite{ref50} model extracts temporal representations of video data to construct a time-series pattern to optimize the multi-instance learning process. MGFN~\cite{mgfn} propose a novel glance and focus network to effectively integrate spatial-temporal information for accurate anomaly detection.In addition, MGFN~\cite{mgfn} propose the Feature Amplifcation Mechanism and a Magnitude Contrastive Loss to enhance the discriminativeness of feature magnitudes for detecting anomalies. NG-MIL~\cite{NG-MIL} encodes diverse normal patterns from noise-free normal videos into prototypes to construct a similarity-based classifier. By combining predictions from classifiers, this approach can refine anomaly scores, reducing training instability from weak labels. CU-Net~\cite{CU-Net} introduces an enhanced framework with a two-stage self-supervised model. This model can generate and iteratively refine pseudo-labels by leveraging the completeness and uncertainty properties of weakly labeled data. HSN~\cite{HSN} propose a Human-Scene Network to learn discriminative representations by capturing both subtle and strong cues in a dissociative manner. Despite their remarkable performance in anomaly detection, these models do not consider factors such as model parameter size and runtime speed in their designs. 
In the there lightweight methods, the work of  Chang et al.\cite{DBLP:journals/tmm/ChangLSFZ22} proposes a lightweight MIL model incorporating a comparative attention module to improve model performance. The method of~\cite{watanabe2022real} introduces a self-attention mechanism to enhance the performance of the lightweight MIL model. BE-WVAD~\cite{ref16} proposes short-input inference modes, which can significantly reduce the required length of input videos and therefore greatly cut down memory and computational costs. Although these two methods achieve considerable advance, there remains substantial room to enhance the design of the model and boost the performance.

\subsection{Implementation Details}
Following the experimental setup in~\cite{DBLP:conf/cvpr/SultaniCS18}, we first divide each video into 32 ($T$=32) video clips. For the model parameters, we have the following settings:
\begin{enumerate}
\itemsep=0pt
\item The number of nodes in the \textit{fully connected} (FC) layer is set to 2048, 64, and 128, respectively. The threshoulds of the Leacky ReLU activation function and dropout function are set to 0.5.
\item  The input features are extracted from the ``mix 5c'' layer of the pre-trained I3D~\cite{DBLP:conf/cvpr/CarreiraZ17} network, and the multi-level temporal correlation attention module is used with a 1×1 Conv1D.
\item  Our model is trained end-to-end using the Adam optimizer~\cite{DBLP:journals/corr/KingmaB14}, with a weight decay coefficient of 0.0005, a batch size of 32, and a total of 200 epochs. The learning rate for both ShanghaiTech and UCF-Crime is set to 0.001. Each batch consists of 32 normal and abnormal instances respectively.
\end{enumerate}

We implement our model using PyTorch. To ensure fairness in the comparison of model performance, we use the same benchmark settings as~\cite{DBLP:conf/cvpr/SultaniCS18,DBLP:journals/tmm/ChangLSFZ22,watanabe2022real}, and report the results of all baselines with the same backbone network as ours.

\subsection{Performance Comparison}
To demonstrate our method's effectiveness, we compare it with existing anomaly detection methods, including general WVAD models and lightweight WVAD methods.

Table~\ref{tab1} presents the frame-level AUC results on the ShanghaiTech dataset. We can see that our method obtains the highest performance of 95.9\% among the lightweight weakly-supervised methods, and surpasses most of general (non-lightweight) methods, whilst maintaining a parameter size of merely 0.14M. 

Table~\ref{tab2} gives the frame-level AUC performance on the UCF dataset. Our method not only achieves the highest AUC of 84.7\% among the lightweight weakly-supervised methods, but also outperforms most non-lightweight weakly-supervised methods, standing at the third place among all general models.

Table~\ref{tab3} gives the frame-level AUC performance on the XD-Violence dataset. Our method not only achieves the highest AUC of 77.3\% among the lightweight weakly-supervised methods, but also performs comparably to most of the non-lightweight weakly-supervised methods.

\begin{table}[]
\caption{Performance comparison of frame-level AUC on ShanghaiTech. For lightweight models, the best result is in red and the second best in blue; For general (Non-lightweight) models, the best result is in bold.}
\label{tab1}
\begin{center}
    
\resizebox{0.48\textwidth}{!}{
\begin{tabular}{c|c|c|c}
\hline
{\color[HTML]{343541} Model type}                                                                            & {\color[HTML]{343541} Method}          & {\color[HTML]{343541} Feature extractor} & {\color[HTML]{343541} AUC~(\%)} \\ \hline
{\color[HTML]{343541} }                                                                                & {\color[HTML]{343541} GCN-Anomaly~\cite{DBLP:conf/cvpr/ZhongLKLLL19}}     & {\color[HTML]{343541} TSN}     & {\color[HTML]{343541} 84.4}   \\
{\color[HTML]{343541} }                                                                                & {\color[HTML]{343541} AR-Net~\cite{DBLP:conf/icmcs/WanFXM20}}          & {\color[HTML]{343541} I3D}     & {\color[HTML]{343541} 91.2}   \\
{\color[HTML]{343541} }                                                                                & {\color[HTML]{343541} CLAWS Net+~\cite{zaheer2022clustering}}      & {\color[HTML]{343541} C3D}     & {\color[HTML]{343541} 89.7}   \\
{\color[HTML]{343541} }                                                                                & {\color[HTML]{343541} MIST~\cite{feng2021mist}}            & {\color[HTML]{343541} C3D}     & {\color[HTML]{343541} 93.1}    \\
{\color[HTML]{343541} }                                                                                & {\color[HTML]{343541} MIST~\cite{feng2021mist}}            & {\color[HTML]{343541} I3D}     & {\color[HTML]{343541} 94.8}    \\
{\color[HTML]{343541} }                                                                                & {\color[HTML]{343541} RTFM~\cite{DBLP:conf/iccv/TianPCSVC21}}            & {\color[HTML]{343541} C3D}     & {\color[HTML]{343541} 91.5}    \\
{\color[HTML]{343541} }                                                                                & {\color[HTML]{343541} RTFM~\cite{DBLP:conf/iccv/TianPCSVC21}}            & {\color[HTML]{343541} I3D}     & {\color[HTML]{343541} 97.2}    \\
{\color[HTML]{343541} }                                                                                & {\color[HTML]{343541} MSL~\cite{DBLP:conf/aaai/00070J22}}             & {\color[HTML]{343541} C3D}     & {\color[HTML]{343541} 94.8}    \\
{\color[HTML]{343541} }                                                                                & {\color[HTML]{343541} MSL~\cite{DBLP:conf/aaai/00070J22}}             & {\color[HTML]{343541} I3D}     & {\color[HTML]{343541} 97.3}    \\
{\color[HTML]{343541} }                                                                                & {\color[HTML]{343541} BN-SVP~\cite{sapkota2022bayesian}}          & {\color[HTML]{343541} C3D}     & {\color[HTML]{343541} 96.0}    \\

{\color[HTML]{343541} }                                                                                & {\color[HTML]{343541} Mu et al.~\cite{DBLP:journals/ipm/MuSWC22}}          & {\color[HTML]{343541} I3D}     & {\color[HTML]{343541} 92.3}    \\

{\color[HTML]{343541} }                                                                                & {\color[HTML]{343541} NTCN-ML~\cite{ref50}}          & {\color[HTML]{343541} I3D+TCN}     & {\color[HTML]{343541} 95.3}    \\

\multirow{-12}{*}{{\color[HTML]{343541} \begin{tabular}[c]{@{}c@{}}General\\ Models\end{tabular}}}     & {\color[HTML]{343541} DAR~\cite{DBLP:journals/tifs/LiuZLK23}}             & {\color[HTML]{343541} I3D}     & {\color[HTML]{343541} {\bf 97.5}}    \\
& {\color[HTML]{343541} NG-MIL~\cite{NG-MIL}}             & {\color[HTML]{343541} I3D}     & {\color[HTML]{343541} 97.4}    \\
& {\color[HTML]{343541} HSN~\cite{HSN}}             & {\color[HTML]{343541} I3D}     & {\color[HTML]{343541} 96.2}    \\

\hline

{\color[HTML]{343541} }                                                                                & {\color[HTML]{343541} Watanabe et al.~\cite{watanabe2022real}} & {\color[HTML]{343541} I3D}     & {\color[HTML]{343541} \textcolor{blue}{95.7}}    \\
{\color[HTML]{343541} }                                                                                & {\color[HTML]{343541} Chang et al.~\cite{DBLP:journals/tmm/ChangLSFZ22}}    & {\color[HTML]{343541} C3D}     & {\color[HTML]{343541} 87.3}    \\
{\color[HTML]{343541} }                                                                                & {\color[HTML]{343541} Chang et al.~\cite{DBLP:journals/tmm/ChangLSFZ22}}    & {\color[HTML]{343541} I3D}     & {\color[HTML]{343541} 92.3}    \\

{\color[HTML]{343541} }                                                                                & {\color[HTML]{343541} BE-WVAD~\cite{ref16}}    & {\color[HTML]{343541} I3D}     & {\color[HTML]{343541} 95.0}    \\

\multirow{-5}{*}{{\color[HTML]{343541} \begin{tabular}[c]{@{}c@{}}Lightweight \\ Models\end{tabular}}} & {\color[HTML]{343541} Light-WVAD (ours)}  & {\color[HTML]{343541} I3D}     & {\color[HTML]{343541} \textcolor{red}{95.9}}    \\ \hline
\end{tabular}
}
\end{center}
\end{table}

Besides comparing the AUC results of different methods on the two datasets, we also compare the number of parameters among all methods, the results are presented in Table~\ref{tab4}. We can see that our method utilizes only 0.14M parameters, which is the lowest among the compared models. Particularly, our method has only less than 1\% of the parameters of the RTMF method, and only half of the parameters of the currently smallest resource-intensive lightweight method~\cite{DBLP:journals/tmm/ChangLSFZ22}.

Combined the comparison results from Table~\ref{tab1} to Table~\ref{tab4}, it is obvious that our method is a lightweight yet good-performance method for weakly supervised video anomaly detection.

\begin{table}[]
\caption{Performance comparison of frame-level AUC on UCF. For lightweight models, the best result is in red and the second best in blue; For general (Non-lightweight) models, the best result is in bold.
}
\label{tab2}
\begin{center}
\resizebox{0.48\textwidth}{!}{
\begin{tabular}{c|c|c|c}
\hline
{\color[HTML]{343541} Model type}                                                                            & {\color[HTML]{343541} Method}          & {\color[HTML]{343541} Feature extractor} & {\color[HTML]{343541} AUC~(\%)} \\ \hline
{\color[HTML]{343541} }                                                                                & {\color[HTML]{343541} Sultani et al.~\cite{DBLP:conf/cvpr/SultaniCS18}}     & {\color[HTML]{343541} TSN}     & {\color[HTML]{343541} 75.4}    \\

{\color[HTML]{343541} }                                                                                & {\color[HTML]{343541} Zhang et al.~\cite{DBLP:conf/icip/ZhangQM19}}     & {\color[HTML]{343541} TSN}     & {\color[HTML]{343541} 78.7}    \\

{\color[HTML]{343541} }                                                                                &  {\color[HTML]{343541} Wu et al.~\cite{DBLP:conf/eccv/Wu0SSSWY20}}                              & {\color[HTML]{343541} I3D}                            & {\color[HTML]{343541} 82.4}                           \\

{\color[HTML]{343541} }                                                                                & {\color[HTML]{343541} GCN-Anomaly~\cite{DBLP:conf/cvpr/ZhongLKLLL19}}     & {\color[HTML]{343541} TSN}     & {\color[HTML]{343541} 82.1}    \\
{\color[HTML]{343541} }                                                                                & {\color[HTML]{343541} CLAWS Net+~\cite{zaheer2022clustering}}      & {\color[HTML]{343541} C3D}     & {\color[HTML]{343541} 83.4}    \\
{\color[HTML]{343541} }                                                                                & {\color[HTML]{343541} MIST~\cite{feng2021mist}}            & {\color[HTML]{343541} C3D}     & {\color[HTML]{343541} 81.4}    \\
{\color[HTML]{343541} }                                                                                & {\color[HTML]{343541} MIST~\cite{feng2021mist}}            & {\color[HTML]{343541} I3D}     & {\color[HTML]{343541} 82.3}    \\
{\color[HTML]{343541} }                                                                                & {\color[HTML]{343541} RTFM~\cite{DBLP:conf/iccv/TianPCSVC21}}            & {\color[HTML]{343541} C3D}     & {\color[HTML]{343541} 83.3}    \\
{\color[HTML]{343541} }                                                                                & {\color[HTML]{343541} RTFM~\cite{DBLP:conf/iccv/TianPCSVC21}}            & {\color[HTML]{343541} I3D}     & {\color[HTML]{343541} 84.3}    \\
{\color[HTML]{343541} }                                                                                & {\color[HTML]{343541} MSL~\cite{DBLP:conf/aaai/00070J22}}             & {\color[HTML]{343541} C3D}     & {\color[HTML]{343541} 82.9}    \\
{\color[HTML]{343541} }                                                                                & {\color[HTML]{343541} MSL~\cite{DBLP:conf/aaai/00070J22}}             & {\color[HTML]{343541} I3D}     & {\color[HTML]{343541} 85.3}    \\

{\color[HTML]{343541} }                                                                                & {\color[HTML]{343541} BN-SVP~\cite{sapkota2022bayesian}}          & {\color[HTML]{343541} C3D}     & {\color[HTML]{343541} 83.4}    \\

{\color[HTML]{343541} }                                                                                & {\color[HTML]{343541} Mu et al.~\cite{DBLP:journals/ipm/MuSWC22}}          & {\color[HTML]{343541} I3D}     & {\color[HTML]{343541} 84.2}    \\

{\color[HTML]{343541} }                                                                                & {\color[HTML]{343541} NTCN-ML~\cite{ref50}}          & {\color[HTML]{343541} I3D+TCN}     & {\color[HTML]{343541} 85.1}    \\

\multirow{-11}{*}{{\color[HTML]{343541} \begin{tabular}[c]{@{}c@{}}General\\ Models\end{tabular}}}     & {\color[HTML]{343541} DAR~\cite{DBLP:journals/tifs/LiuZLK23}}             & {\color[HTML]{343541} I3D}     & {\color[HTML]{343541} 85.2}    \\
& {\color[HTML]{343541} CU-Net~\cite{CU-Net}}             & {\color[HTML]{343541} I3D}     & {\color[HTML]{343541} 86.2}    \\ 
& {\color[HTML]{343541} MGFN~\cite{mgfn}}             & {\color[HTML]{343541} I3D}     & {\color[HTML]{343541} {\bf 87.0}}    \\ 
& {\color[HTML]{343541} NG-MIL~\cite{NG-MIL}}             & {\color[HTML]{343541} I3D}     & {\color[HTML]{343541} 85.6}    \\ 
& {\color[HTML]{343541} HSN~\cite{HSN}}             & {\color[HTML]{343541} I3D}     & {\color[HTML]{343541} 85.5}    \\ 

\hline
{\color[HTML]{343541} }

& {\color[HTML]{343541} Watanabe et al.~\cite{watanabe2022real}} & {\color[HTML]{343541} I3D}     & {\color[HTML]{343541} \textcolor{red}{84.7}}    \\
{\color[HTML]{343541} }                                                                                & {\color[HTML]{343541} Chang et al.~\cite{DBLP:journals/tmm/ChangLSFZ22}}    & {\color[HTML]{343541} C3D}     & {\color[HTML]{343541} 83.4}    \\
{\color[HTML]{343541} }                                                                                & {\color[HTML]{343541} Chang et al.~\cite{DBLP:journals/tmm/ChangLSFZ22}}    & {\color[HTML]{343541} I3D}     & {\color[HTML]{343541} \textcolor{blue}{84.6}}    \\

{\color[HTML]{343541} }                                                                                & {\color[HTML]{343541} BE-WVAD.~\cite{ref16}}    & {\color[HTML]{343541} I3D}     & {\color[HTML]{343541} 84.1}    \\

\multirow{-4}{*}{{\color[HTML]{343541} \begin{tabular}[c]{@{}c@{}}Lightweight \\ Models\end{tabular}}} & {\color[HTML]{343541} Light-WVAD (ours)}  & {\color[HTML]{343541} I3D}     & {\color[HTML]{343541} \textcolor{red}{84.7}}    \\ \hline
\end{tabular}
}
\end{center}
\end{table}

\begin{table}[]
\begin{center}
\caption{Performance comparison of frame-level AP on XD-Violence. For lightweight models, the best result is in red and the second best in blue; For general (Non-lightweight) models, the best result is in bold.}
\resizebox{0.48\textwidth}{!}{
\label{tab3}
\begin{tabular}{c|c|c|c}
\hline
Model type                                                                    & Method         & Feature extractor & AP(\%) \\ \hline
\multirow{10}{*}{\begin{tabular}[c]{@{}c@{}}General\\ Models\end{tabular}}    & Sultani et al.~\cite{DBLP:conf/cvpr/SultaniCS18} & C3D               & 73.2   \\
& Wu et al.~\cite{DBLP:conf/eccv/Wu0SSSWY20}      & I3D               & 75.4   \\
& RTFM~\cite{DBLP:conf/iccv/TianPCSVC21}           & C3D               & 75.9   \\
& RTFM~\cite{DBLP:conf/iccv/TianPCSVC21}           & I3D               & 77.8   \\
& MSL~\cite{DBLP:conf/aaai/00070J22}            & C3D               & 75.5   \\
& MSL~\cite{DBLP:conf/aaai/00070J22}            & I3D               & 78.3   \\
& DAR~\cite{DBLP:journals/tifs/LiuZLK23}            & I3D               & 78.9   \\
& MGFN~\cite{mgfn}           & I3D               & {\bf 79.2}   \\
& NG-MIL~\cite{NG-MIL}         & I3D               & 78.5   \\
& CU-Net~\cite{CU-Net}         & I3D               & 78.7   \\ \hline
\multirow{4}{*}{\begin{tabular}[c]{@{}c@{}}Lightweight\\ Models\end{tabular}} 
& Chang et al.~\cite{DBLP:journals/tmm/ChangLSFZ22}   & I3D+flow          & 71.5   \\
& Chang et al.~\cite{DBLP:journals/tmm/ChangLSFZ22}   & I3D               & \textcolor{blue}{76.9}   \\
& BE-WVAD~\cite{ref16}        & I3D               & 74.9   \\
& Light-WVAD(ours)       & I3D               & \textcolor{red}{77.3}   \\ \hline
\end{tabular}
}
\end{center}
\end{table}

\begin{table}[]
\begin{center}
\caption{Model size comparison among weakly supervised methods. The best result is in red and the second best is in blue.}
\label{tab4}
\begin{tabular}{c|c}
\hline
Method          & \#Parameters~(M) \\ \hline
Sultani et al.~\cite{DBLP:conf/cvpr/SultaniCS18}  & 2.11                    \\
Wu et al.~\cite{DBLP:conf/eccv/Wu0SSSWY20}       & 0.76                    \\
RTFM~\cite{DBLP:conf/iccv/TianPCSVC21}            & 24.72                   \\
Mu et al.~\cite{DBLP:journals/ipm/MuSWC22}       & 13.20                   \\
BE-WVAD~\cite{ref16} & 2.49                    \\
Watanabe et al.~\cite{watanabe2022real} & 0.33                    \\
Chang et al.~\cite{DBLP:journals/tmm/ChangLSFZ22}    & \color{blue}{0.26}                    \\
Light-WVAD (ours)   & \color{red}{0.14}                    \\ \hline
\end{tabular}
\end{center}
\end{table}

\subsection{Ablation Studies}
\textbf{Effect of major modules in Light-WVAD.}
To evaluate the effectiveness of the major modules in our method, we conduct ablation experiment on the ShanghaiTech dataset. The results are presented in Table~\ref{tab5}. Here, the baseline model is a simple network consisting of fully connected layers, with an AUC of 93.8\%. The results in Table~\ref{tab5} show that employing MTA, AIS, and the antagonistic loss (A-Loss in short) to the baseline individually can obviously improve the model's performance. Furthermore, our proposed lightweight HFC structure does not have negative impact on the model's performance. In summary, by combining MTA, HFC, AIS, and A-Loss into our method, a 2.1\% performance improvement on the ShanghaiTech dataset is achieved.

\begin{table}[]
\caption{Ablation study on ShanghaiTech.}
\begin{center}
\label{tab5}
\resizebox{0.48\textwidth}{!}{
\begin{tabular}{ccccc|c}
\hline
Baseline & MTA & HFC & AIS & A-Loss & AUC (\%)-SH \\ \hline
\checkmark        &    &     &     &        & 93.8       \\
\checkmark        & \checkmark   &     &     &        & 94.8       \\
\checkmark        &     & \checkmark   &     &        & 93.9       \\
\checkmark        & \checkmark   & \checkmark   &     &        & 94.9       \\ \hline
\checkmark        &     &     & \checkmark   &        & 95.0       \\
\checkmark        &     & \checkmark   & \checkmark   &        & 95.1       \\
\checkmark        & \checkmark   &     & \checkmark   &        & 95.4       \\
\checkmark        & \checkmark   & \checkmark   & \checkmark   &        & 95.4       \\ \hline
\checkmark        &     &     &     & \checkmark      & 95.4       \\
\checkmark        & \checkmark   & \checkmark   & \checkmark   & \checkmark      & \color{red}{95.9}       \\ \hline
\end{tabular}
}
\end{center}
\end{table}

\textbf{Effect of parameter $k$ in MTA.} MTA is to capture multi-level temporal correlations of consecutive instances by integrating inter-instance feature relationships across multiple time intervals. As described in Section~\ref{sec:mta}, MTA has a hyperparameter \textit{$k$}, which represents the maximum number of consecutive \textit{$k$} instances, from which MTA can extract temporal correlation information. The value of \textit{$k$} affects the model's performance. We change the value of $k$ from 3 to 15 with a stepsize of 2, and report the performance results in Table~\ref{tab6}. We can see that when \textit{$k$} is set to 5, the model achieves the best performance. By analyzing the performance change when setting different values of \textit{k} for MTA, we can see that abnormal events typically have a relatively short duration. When \textit{$k$} is too large, the proportion of normal instances is too high, which may dampen the influence of abnormal instances so that MTA fails to provide discriminative temporal correlation information. On the other hand, a too small \textit{$k$} may result in insufficient coverage of anomalous instances, so that MTA cannot obtain effective temporal correlation information by enough inter-instance information.

\begin{table}[]
\caption{Ablation study on parameter $k$ in MTA on ShanghaiTech. }
\label{tab6}
\begin{center}
\begin{tabular}{c|c|c}

\hline
Model                         & Hyperparameter $k$ & AUC~(\%)  \\ \hline
Baesline                      & -                & 93.8 \\ \hline
\multirow{7}{*}{Baseline+MTA} & 3                & 94.4 \\
                              & 5                & \textcolor{red}{94.8} \\
                              & 7                & 94.5 \\
                              & 9                & 93.9 \\
                              & 11               & 93.3 \\
                              & 13               & 93.3 \\
                              & 15               & 93.0 \\ \hline
\end{tabular}
\end{center}
\end{table}

\textbf{Effect of the antagonistic loss function.} %The sparsity loss presupposes that the mean anomaly score of abnormal instances in abnormal videos should approximate zero. Nonetheless, in the practical model training scenario, this assumption does not stand due to the brief duration of the employed abnormal videos and the non-trivial proportion of abnormal instances. Hence, we suggest an antagonistic loss that is more applicable to the prevailing experimental configurations. 
To further verify the advantage of the proposed antagonistic loss, we compare the performance of three configurations: not using the sparsity loss, using the sparsity loss, and using our antagonistic loss (our method) on the ShanghaiTeah dataset, the results are presented in Table~\ref{tab7}. We can see that the antagonistic loss can obviously improve model performance, whereas the sparsity loss results in a decrease in model performance.

\begin{figure}[]
	\centering
		\includegraphics[scale=0.28]{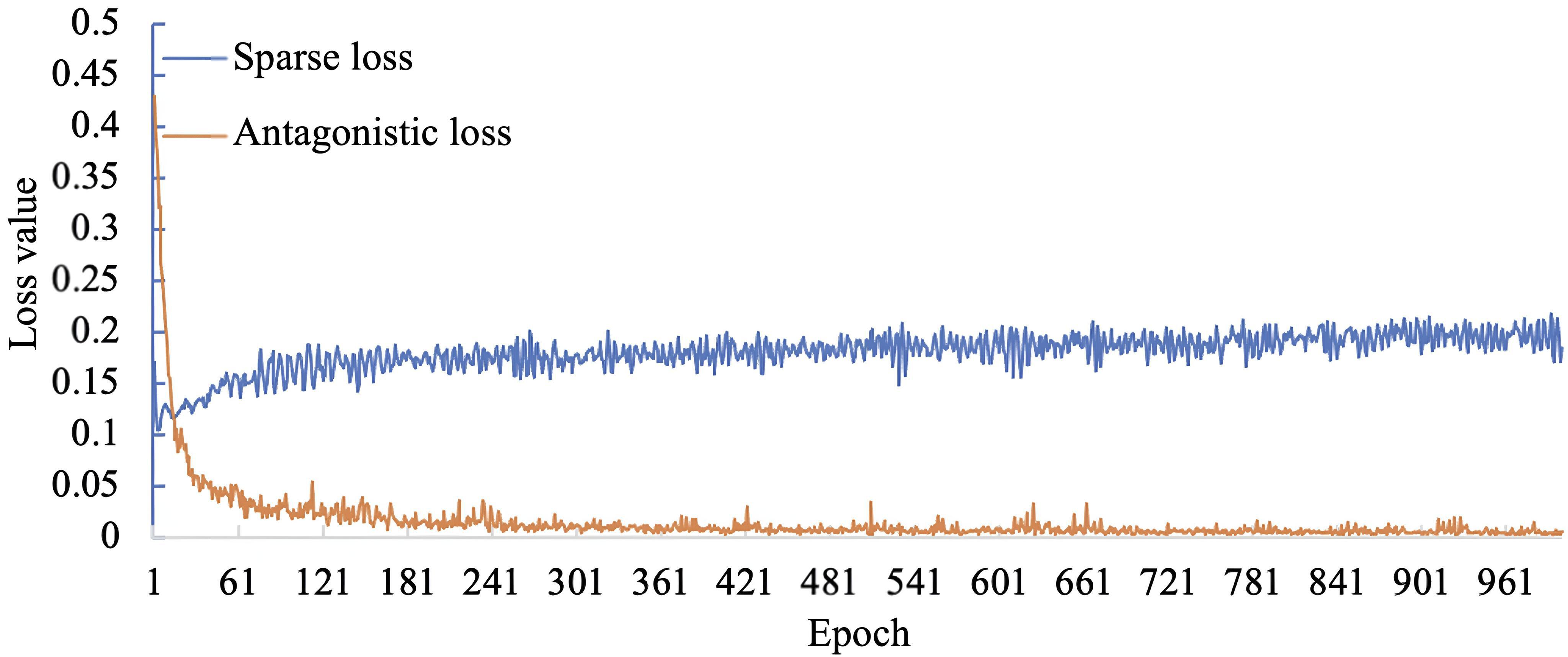}
	\caption{The loss curves in model training when using (a) the psarsity loss and (b) our antagomostic loss.}  %\textcolor{red}{(a) represents the traditional fully connected layer, (b) represents hourglass-shaped fully connected layer}.}
	\label{FIG:5}
\end{figure}

To delve deeper into why the sparsity loss causes model performance degradation, we train models employing the antagonistic loss and the sparsity loss respectively on the ShanghaiTeah dataset, and illustrate their loss curves in Fig.~\ref{FIG:5},  from which we can see a rapid drop of the antagonistic loss as the training goes, conforming to our expectation, whereas the sparsity loss curve exhibits an upward trend, which suggests a gradual increase in the average anomaly score of abnormal videos during the training. This indicates the unsuitability of the sparsity loss for model training, and further indicates that the proportion of abnormal instances in abnormal videos should not be ignored.

\begin{table}[]
\caption{Ablation study on loss functions in our method on ShanghaiTech. }
\label{tab7}
\begin{center}
\begin{tabular}{c|c}
\hline
Model                                                                            & AUC (\%)-SH \\ \hline
\begin{tabular}[c]{@{}c@{}}Ours\\ ($loss_{AIS}+loss_{smooth}$)\end{tabular}              & 95.4       \\ \hline
\begin{tabular}[c]{@{}c@{}}Ours\\ ($loss_{AIS}+loss_{smooth}+loss_{sparse}$)\end{tabular}   & 94.8       \\ \hline
\begin{tabular}[c]{@{}c@{}}Ours\\($loss_{AIS}+loss_{smooth}+loss_{antagonistic}$)\end{tabular} & \textcolor{red}{95.9}       \\ \hline
\end{tabular}
\end{center}
\end{table}

\subsection{Visual Analysis}
\begin{figure}[t!]
	\centering
		\includegraphics[scale=0.105]{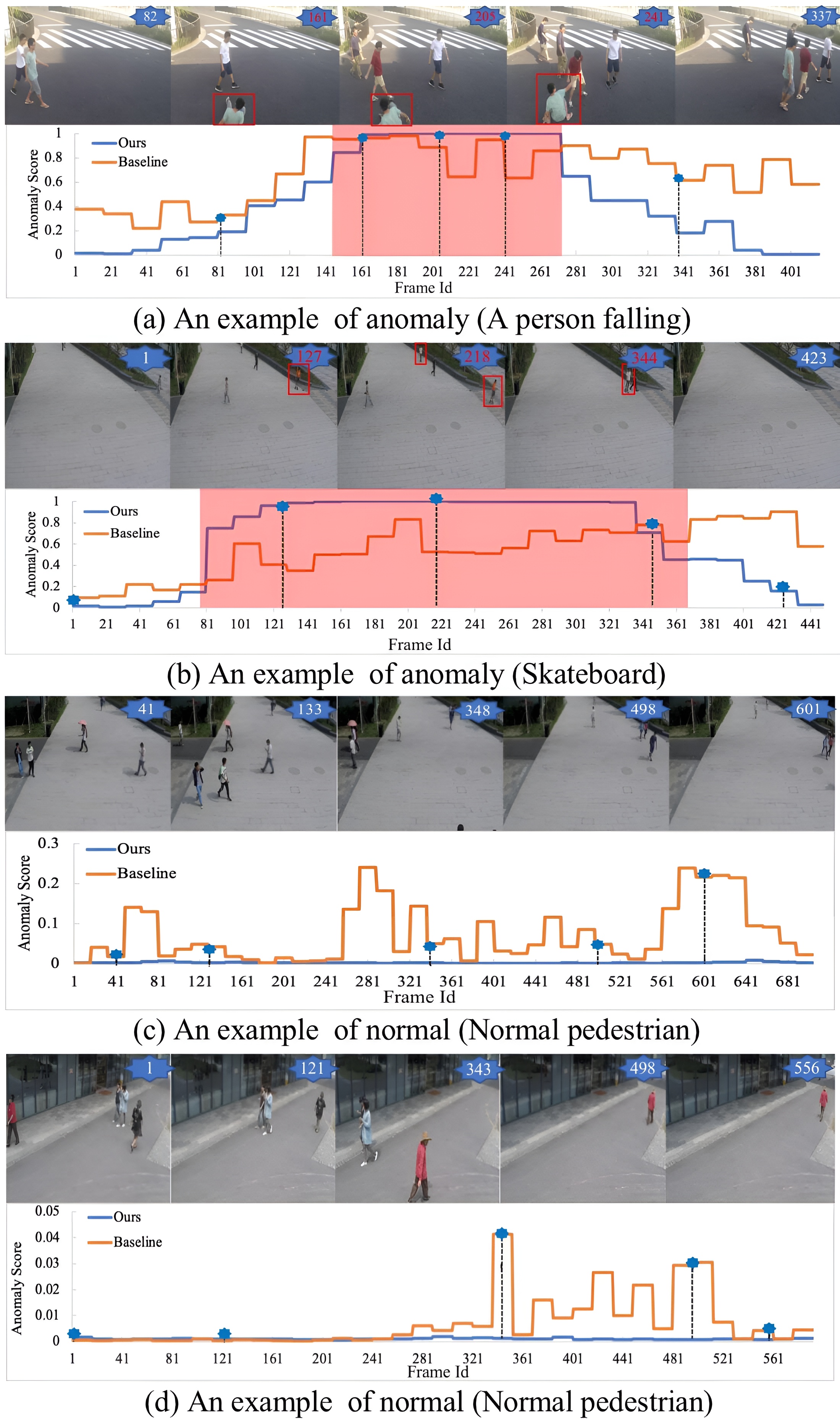}
	\caption{Visualization of test results of our method Light-WVAD and the baseline model on the ShanghaiTech dataset. The pink area denotes the time period that anomalies occur in the video, and the numerical labels on the line correspond to the labels in the video frames. In the anomalous video screenshots, the numbers are highlighted in red, and the anomalous objects are marked with red boxes in the frames.}
	\label{FIG:6}
\end{figure}

\begin{figure}[t!]
    \centering
        \includegraphics[scale=0.11]{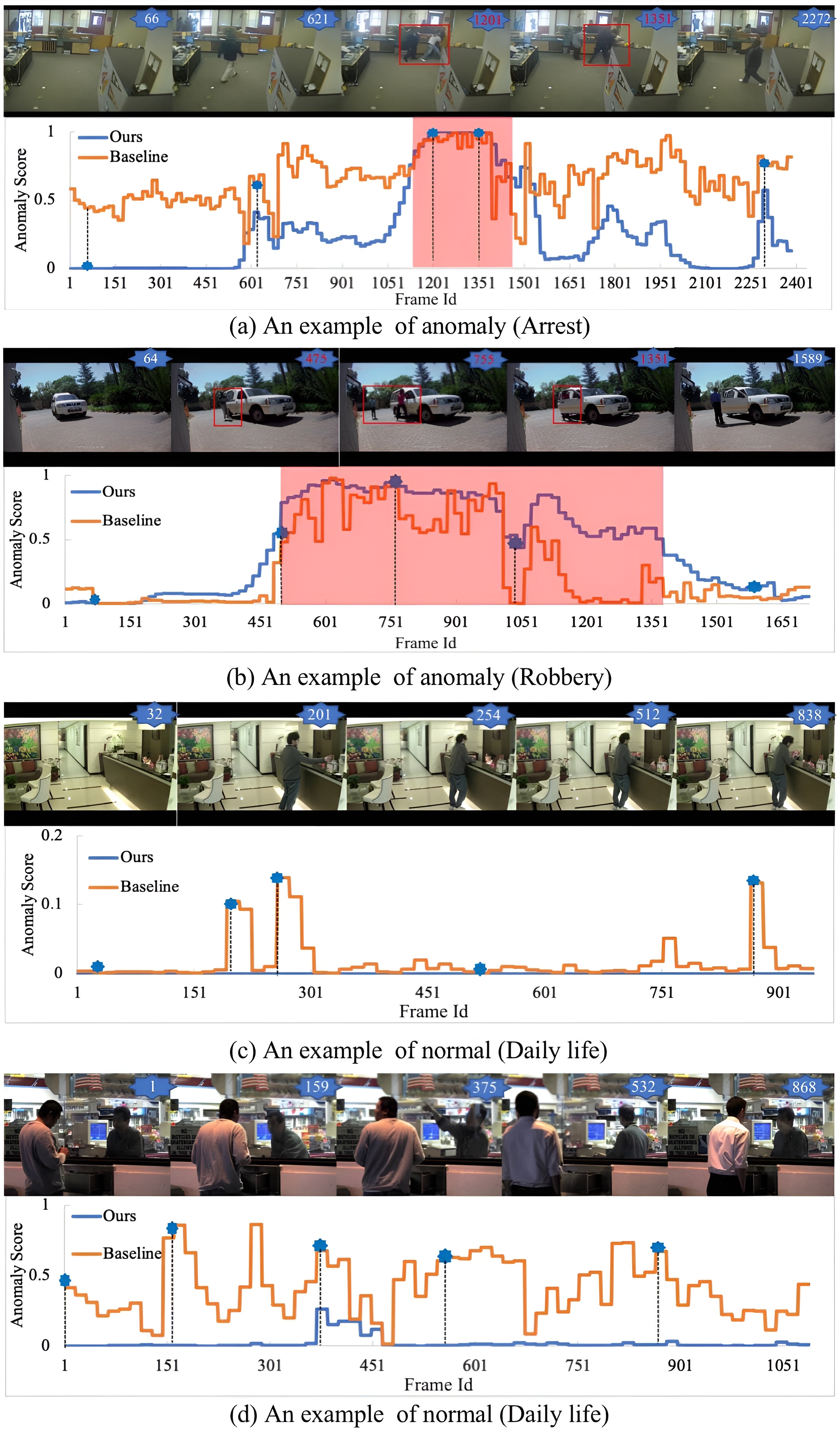}
    \caption{Visualization of test results of our method Light-WVAD and the baseline model on the UCF-Crime dataset. The pink area denotes the time period that anomalies occur in the video, and the numerical labels on the line correspond to the labels in the video frames. In the anomalous video screenshots, the numbers are highlighted in red, and the anomalous objects are marked with red boxes in the frames.}
    \label{FIG:7}
\end{figure}

\textbf{Visualization of test results.}
Here, we visualize the test results of our method and the baseline model on the ShanghaiTech and UCF-Crime datasets in Fig.~\ref{FIG:6} and Fig.~\ref{FIG:7} respectively, to further demonstrate the performance of our method.

In Fig.~\ref{FIG:6}(a), the anomalous event is a man falling to the ground, and in Fig.~\ref{FIG:6}(b) the anomalous event is someone playing skateboard on pavement. However, in Fig.~\ref{FIG:6}(c) and Fig.~\ref{FIG:6}(d), there is no anomaly. 

As illustrated in Fig.~\ref{FIG:6}(a) and Fig.~\ref{FIG:6}(b), our method can accurately identify the anomalous frames (e.g. No.~161, 205 and 241 in Fig.~\ref{FIG:6}(a), and No.~127, 218 and 344 in Fig.~\ref{FIG:6}(b)), i.e., assigning very high anomaly scores to these anomalous frames, while the baseline model identifys some normal frames as anomaly (e.g. Frame No.~337 in Fig.~\ref{FIG:6}(a) and Frame No.~423 in Fig.~\ref{FIG:6}(b)), and assigns low scores to some anomalies (e.g. Frame 127 in  Fig.~\ref{FIG:6}(b)). 
This indicates that our method has stronger capability of anomaly detection, and is more accurate in detecting the starting and ending of anomalies than the baseline model. Furthermore, Fig.~\ref{FIG:6}(c) and Fig.~\ref{FIG:6}(d) show that our method is more stable on normal video detection than the baseline model by consistently maintaining lower anomaly scores for normal frames. 

In Fig.~\ref{FIG:7}(a) and Fig.~\ref{FIG:7}(b), the anomaly is an arrest action and a robbery event, respectively, while in Fig.~\ref{FIG:7}(c) and Fig.~\ref{FIG:7}(d), there is no anomaly. 

The visualization results in Fig.~\ref{FIG:7}(a) indicate that our method can accurately identify the anomalous event of police performing an arrest action. Nonetheless, we also observe incorrect responses at Frame No.~621 and No.~2272 in the normal screenshots, which are attributable to the drastic movements of individuals in the scene. Thus, we guess that our model faces challenge in delineating certain normal-anomaly boundaries, primarily due to the absence of detailed annotations. In Fig.~\ref{FIG:7}(b), the anomaly corresponds to a gunman robbing a car, and a noticeable score fluctuation occurs at Frame No.~1351 in the video, due to the gunman being obscured by the car door, preventing the model from recognizing the anomaly. Nevertheless, outside the scope of the car door's obstruction, our model can still successfully detect the anomaly. Ultimately, as per the visualization results in Fig.~\ref{FIG:7}(c) and Fig.~\ref{FIG:7}(d), our model still exhibits obvious stability on normal videos.

\begin{figure}[]
	\centering
		\includegraphics[scale=0.185]{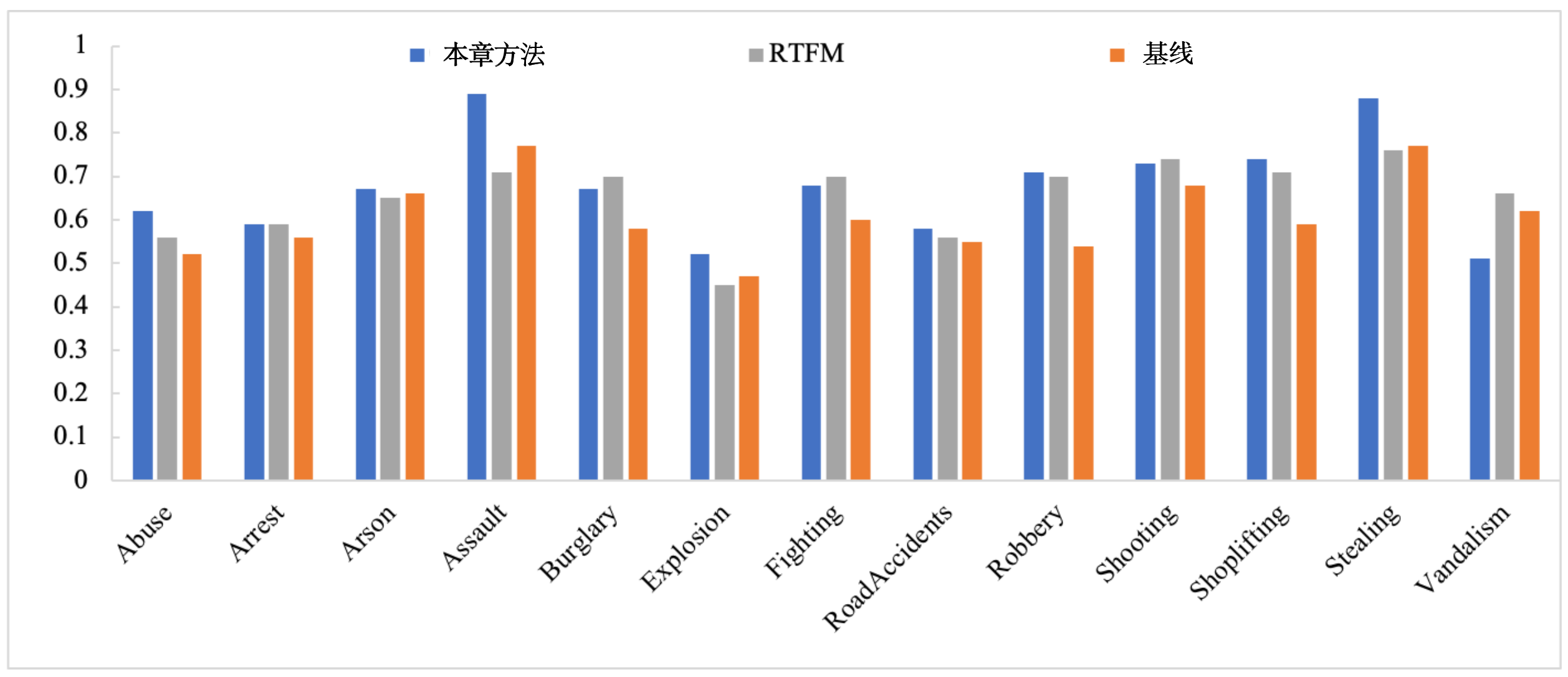}
	\caption{AUC results of three methods on different anomalous classes in the UCF-Crime dataset.}
	\label{FIG:8}
\end{figure}

\textbf{Detection performance of each anomalous class in the UCF-Crime dataset.} Fig.~\ref{FIG:8} presents the AUC of our method on each anomalous class in the UCF-Crime dataset. Compared to RTFM and the baseline, our method yields superior or equivalent detection accuracy across 11 anomaly classes. Specifically, AUC is improved over 12\% for the ``Assault'' and ``Stealing'' classes. Typically, without long-term analysis of object and human motion, detecting these two types of anomalies is very difficult. Nevertheless, our method achieves high detection accuracy, indicating that our MTA module is effective. Similar to the other methods, our model exhibits suboptimal performance on the ``Explosion'', ``Road Accidents'', ``Vandalism'', and ``Abuse'' classes. For the sudden anomalies without enough warning signals, they remains a challenge for video anomaly detection. %Detecting instantaneous anomalies without \textcolor{red}{any prior warning} remains a challenge for video anomaly detection.

\section{Conclusion}\label{sec:conclusion}
This study develops a lightweight weakly-supervised video anomaly detection method (Light-WVAD) that can effectively addresses the uncertainty and high-parameter issues associated with the existing WVAD methods. Compared with existing lightweight models, Light-WVAD has the smallest number of parameters and the best performance. However, predicting sudden anomalies without warning signals is still a serious challenge for current video anomaly detection methods. And the predicted results on different anomalous classes in the UCF-Crime dataset also show that our model is not effective enough in detecting sudden anomaly classes. In the future, we plan to employ multi-modal techniques to enhance the model's understanding of video content, and thus develop more effective video anomaly detection models.

%\appendix

\printcredits

\section*{Acknowledgement}
This work was supported in part by National Natural Science Foundation of China (No. U1936205, 
No. 6217230
0), National Key R\&D Program of China (No. 2021YFC330030
0), Open Research Projects of Zhejiang Lab (No. 2021KH0A
B04), and the Fundamental Research Funds for the Central Universities (No. ZD-21-202101). Finally, we would like to thank Ang Li and Jun Yan for their corrections to the writing of this paper.

\bio{Wangyang}
\textbf{Yang Wang} received the M.S. degree in Computer technology from the Nanchang University
, Nanchang, China, in 2017. He is currently working toward the Ph.D. degree in Computer Science with the Tongji University, Shanghai, China. His research interests include anomaly detection, object detection and deep learning. 
\endbio

\bio{Zhoujiaogen}
\textbf{Jiaogen Zhou} received the bachelor's degree from University of Chinese Academy of Sciences, China and the PhD degree from Wuhan University, China. His research interests include Image classification, object detection, animal pose estimation and machine learning.
\endbio
\bio{guanjihong}
\textbf{JihongGuan}Jihong Guan is now a professor of Department of Computer Science \& Technology, Tongji University, Shanghai, China. She received
his Bachelor’s degree from Huazhong Normal University in 1991, her Master’s degree from Wuhan Technical University of Surveying and Mapping
(merged into Wuhan University since Aug. 2000) in 1991, and her Ph.D. from Wuhan University in 2002. Before joining Tongji University, she served in the Department of Computer, Wuhan Technical University of Surveying and Mapping from 1991 to 1997, as an assistant professor and an associate professor (since August 2000) respectively. She was an associate professor (Aug. 2000-Oct. 2003) and a professor (Since Nov. 2003) in the School of Computer, Wuhan University. Her research interests include spatial databases, artificial intelligence and bioinformatics. She has published more than 200 papers in domestic and international journals and conferences.
\endbio
\end{document}